\documentclass[runningheads]{llncs}
\usepackage{soul}
\usepackage[dvipsnames]{xcolor}
\usepackage{graphicx}
\usepackage{babel}
\usepackage{times}
\usepackage{amsmath, amsfonts, amssymb}
\usepackage{tikz}
\usetikzlibrary{arrows.meta}
\usepackage{caption}
\usepackage{subcaption}
\usepackage{booktabs}
\usepackage{multirow}
\usepackage{caption}
\usepackage{subcaption}

\sethlcolor{green}
\usepackage{tabularx}
\newcolumntype{C}{>{\centering\arraybackslash}X}

\begin{document}
\title{Maximal Independent Vertex Set applied to Graph Pooling}
%
%
\author{Stevan Stanovic\inst{1}\orcidID{0000-0001-9656-2080} \and
Benoit Gaüzère\inst{2}\orcidID{0000-0001-9980-2641} \and
Luc Brun\inst{1}\orcidID{0000-0002-1658-0527}}
\authorrunning{S. Stanovic et al.}
%
\institute{Normandie Univ, ENSICAEN, CNRS, UNICAEN, GREYC UMR 6072, 14000 Caen, France \\ \email{\{stevan.stanovic, luc.brun\}@ensicaen.fr} \and
Normandie Univ, INSA de Rouen, Univ. rouen, Univ. Le Havre, LITIS EA 4108,
76800 Saint-Étienne-du-Rouvray, France\\ \email{benoit.gauzere@insa-rouen.fr}}
\maketitle              
\begin{abstract}
    Convolutional neural networks (CNN) have enabled major advances in image classification through convolution and pooling. In particular, image pooling transforms a connected discrete grid into a reduced grid with the same connectivity and allows reduction functions to take into account all the pixels of an image. However, a pooling satisfying such properties does not exist for graphs. Indeed, some methods are based on a vertex selection step  which  induces an important loss of information. Other methods learn a fuzzy clustering of vertex sets which induces almost complete reduced graphs. We propose to overcome both problems using a new pooling method, named MIVSPool. This method is based on a selection of vertices called surviving vertices using a Maximal Independent Vertex Set (MIVS) and an assignment of the remaining vertices to the survivors. Consequently, our method does not discard any vertex information nor artificially increase the density of the graph. Experimental results show an increase in accuracy for graph classification on various standard datasets.

    \keywords{Graph Neural Networks \and Graph Pooling \and Graph Classification \and Maximal Independant Vertex Set.}
\end{abstract}
\renewcommand{\thefootnote}{}
    \section{Introduction}
        Convolutional neural networks (CNN)\footnotetext{The work reported in this paper was supported by French ANR grant \#ANR-21-CE23-0025 CoDeGNN} have enabled major advances in image classification. An image can be defined as a connected discrete grid and this property enables to define efficient convolution and pooling operations. Nevertheless, social networks, molecules or traffic infrastructures are not represented by a grid but by graphs. Convolution and pooling have been adapted to these structured data into Graph Neural Networks (GNN)~\cite{scarselli2008graph}.
        The adaptation of convolution to graphs can be performed by using learned aggregation functions which combine the value of each vertex with the ones of its neighborhood~\cite{hamilton2017inductive,kipf2017semi}.
        For graph pooling, the adaptation is mainly performed by selecting vertices~\cite{bianchi2020hierarchical,gao2019graph,lee2019self,zhang2021hierarchical} or by learning a fuzzy clustering~\cite{ying2018hierarchical}.
        
\begin{figure}[t!]
                \centering
                \includegraphics[width=\textwidth]{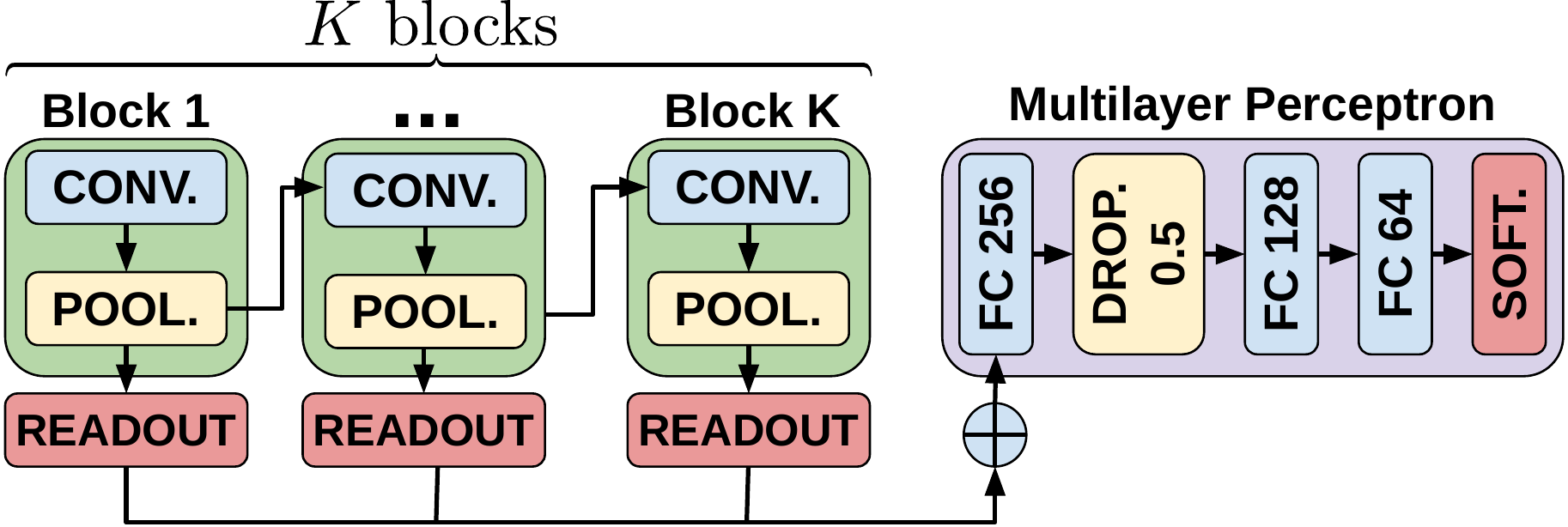}
                \caption{General architecture of our GNN. Each block is composed of a convolution layer followed by a pooling layer. Features learned after each block are combined to have several levels of description of the graph.}
                \label{fig:model}
        \end{figure}

\begin{figure}[t!]
        \centering
        \includegraphics[width=\textwidth]{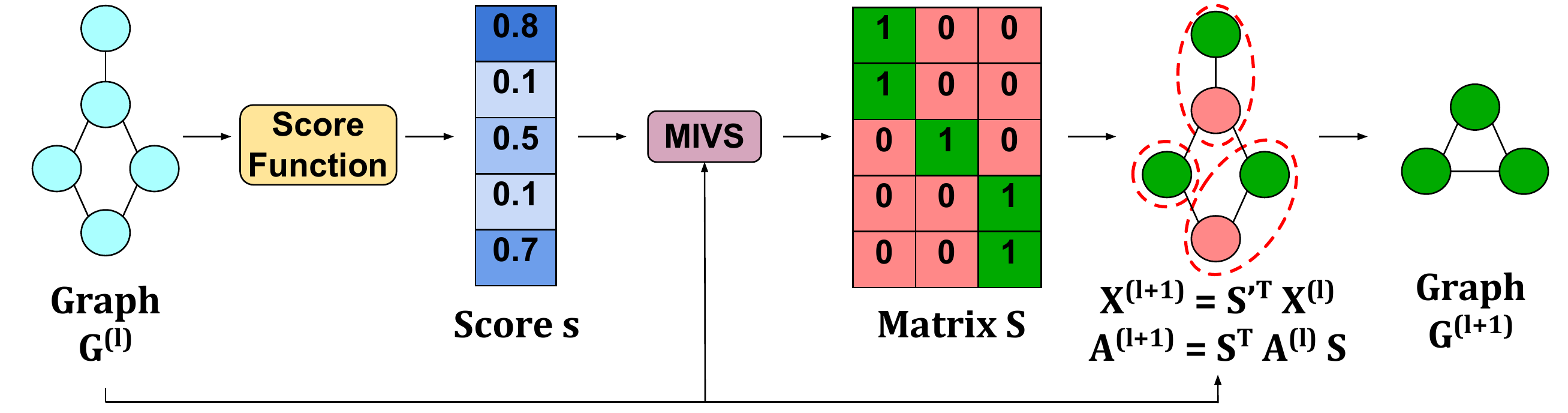}
                \caption{Proposed graph pooling. Each node is associated to a score (Section~\ref{subsec:scoring}). Based on this score, a MIVS is computed from which a reduction matrix $S$ is derived. Applying $S$ to both feature and structure leads to a reduced graph $G^{(l+1)}$. }
                
            \label{fig:pooling}
        \end{figure}

        In this paper, we detail graph convolution and pooling as well as the construction of  the reduced graph in Section~\ref{sec:relatedWork}. Section~\ref{sec:proposedMethod} presents our method (Fig.~\ref{fig:pooling}) and explains its differences from other pooling methods. We present in Section 4 our experiments where we study different heuristics to select surviving vertices and the mean complexity of our pooling algorithm.  We finally compare our method to other pooling strategies on standard datasets using an unified architecture (Fig.~\ref{fig:model}) in Section~\ref{sec:xp}.
    
    \section{Related Work}
    \label{sec:relatedWork}
        
        GNNs are neural networks applied to graphs. Inspired by CNNs, they are like the latter based on convolution and pooling operations. Let us note that convolution operation should be permutation equivariant. Intuitively, this condition states that a convolution operation may be permuted with a permutation of the graph's nodes.
        
        Convolution operations diffuse the information by a message passing mechanism which allows to learn a representation for each node by aggregating the information of their \textit{n}-hop neighborhood. Optimized according to a particular task, the resulting nodes' features can be used as input of a node classifier or regressor. According to~\cite{balcilar2020analyzing}, current aggregation functions correspond mainly to low-pass filters.

        \subsection{Graph Pooling}
        \label{subsec:state_art_pooling}
        
        Considering graph level tasks like graph classification, we need to compute a graph representation as a fixed sized vector by summing node's representation. This operation is called global pooling. Global pooling realizes an aggregation of all the vertices of a graph and summarizes it in a single vector. Such aggregation must be permutation invariant and is usually performed using basic  operators like a sum, mean, maximum or more complex ones~\cite{zhang2018end}.  
        
        However, global pooling performed on all the vertices of a graph  leads to sum up an unbounded amount of information (proportional to the graph's size) into a fixed size vector and thus potentially induces a large amount of information loss. Many authors~\cite{lee2019self,ying2018hierarchical} have proposed to decompose GNNs in several steps alternating convolution and pooling to reduce the graph size while averaging vertices values. Such methods are called hierarchical pooling. The choice of the number of vertices for the reduced graph can be fixed or adapted according to the original graph's size with a ratio. The hierarchical pooling, like the convolution operation should be permutation equivariant. 
        
        \textbf{Notations.}
        For any pooling layer $l$, we consider a graph $\mathcal{G}^{(l)} = (\mathcal{V}^{(l)}, \mathcal{E}^{(l)})$ where $\mathcal{V}^{(l)}$ and $\mathcal{E}^{(l)}$ are respectively the set of vertices and the set of edges of the graph. Let $n_{l} = |\mathcal{V}^{(l)}|$, we can define $\textbf{A}^{(l)} \in \mathbb{R}^{n_{l} \times n_{l}}$ the adjacency matrix associated to the graph $\mathcal{G}^{(l)}$ where $\mathbf{A}^{(l)}_{ij} = 1$ if it exists an edge between the vertices $i$ and $j$, $0$ otherwise. We also note $\textbf{X}^{(l)} \in \mathbb{R}^{n_{l} \times f_{l}}$ the feature matrix of the graph $\mathcal{G}^{(l)}$ where $f_{l}$ is the dimension of nodes' attributes.
        
        \textbf{Construction of the set of surviving vertices.}
        It exists multiple methods to reduce the size of a graph within the GNN framework. However, most of these methods lead to the construction of a reduction matrix $S^{(l)} \in \mathbb{R}^{n_l \times n_{l+1}}$  where $n_l$ and $n_{l+1}$ are respectively the sizes of the original and the reduced graph. This matrix is used to define the attributes and the adjacency matrix of the reduced graph. Each surviving vertex $i$ contributing to its own cluster, we suppose that $S^{(l)}_{i,i}=1$.
        
        \textbf{Construction of attributes and adjacency matrix.}
        The reduction matrix $S^{(l)}$ is the basis of the construction of the reduced graph. Based on $S^{(l)}$,  the following two equations allow this construction:
        \begin{equation}
            X^{(l+1)} = S^{(l)\top} X^{(l)}\label{eq:2}
        \end{equation}
        This last equation defines the attribute of each surviving vertex $v_i$ as a weighted sum of the attributes of the  vertices $v_j$ of $G^{(l)}$ such that  $S^{(l)}_{ji}\neq 0$.
        \begin{equation}
            A^{(l+1)} = S^{(l)\top} A^{(l)} S^{(l)}\label{eq:3}
        \end{equation}
        Equation~\eqref{eq:3} can be rewritten as follows for any pair of surviving vertex $(i,j)$:
        \begin{equation}
            (S^{(l)\top} A^{(l)} S^{(l)})_{i,j} = \sum^{n_{l}}_{k,l} A^{(l)}_{k,l} S^{(l)}_{k,i} S^{(l)}_{l,j}
        \label{eq:4}
        \end{equation}
        Two surviving vertices are therefore adjacent in the reduced graph if they are adjacent in the initial graph $(A^{(l)}_{i,j} = S^{(l)}_{i,i} = S^{(l)}_{j,j} = 1)$. Moreover, surviving vertices $i$ and $j$ are adjacent in the reduced graph if it exists a pair of non-surviving adjacent vertices $(k,l)$ assigned respectively to $i$ and $j$ $(A^{(l)}_{k,l} = S^{(l)}_{k,i} = S^{(l)}_{l,j} = 1)$.
        
        \textbf{Families of methods.} Pooling methods can be divided in two families. First family consists of methods based on the selection of  surviving vertices on a given criteria. This criteria can be  the result of a combinatorial algorithm~\cite{bianchi2020hierarchical} or a learning step like in Top-\textit{k} methods~\cite{gao2019graph,lee2019self}.
        The second family regroups methods based on a node's clustering as in DiffPool~\cite{ying2018hierarchical}. Each cluster is associated to a surviving vertex.
        
        Methods of the second group use a fixed number of clusters. Hence, learning $S^{(l)}$ does not allow these methods to take into account the variable size and topology of the graphs. Moreover, due to training, such methods  produce dense matrices $S^{(l)}$ with few nonzero values. Equation~\eqref{eq:4} shows that the matrix $A^{(l+1)}$ is then dense and the corresponding graph has a density close to $1$ (i.e. will be a complete graph or almost complete graph). Consequently, the structure of the graph is not respected. We say that $S^{(l)}$ is \textit{complete}.
        
        For Top-\textit{k} methods, we define the reduction matrix $S^{(l)}$ as the restriction of the identity matrix $I_{n_l}$ to the column indices $idx$ corresponding to the surviving vertices:
        \begin{equation}
            S^{(l)} = [I_{n_l}]_{:, idx}\label{eq:5}
        \end{equation}
        Given a surviving vertex $i$ in the original graph, we have thus $S^{(l)}_{i,\phi(i)} = 1$, where $\phi(i)$ denotes the column index of $i$ in the reduced matrix $S^{(l)}$. All other entries of $S^{(l)}$ are set to $0$. The matrix $S^{(l)}$ is called \textit{selective} since it selects the attributes of the surviving vertices and removes those of non-surviving vertices. Moreover, in this case, rows of  $S^{(l)}$  corresponding to non-surviving vertices are equal to $0$ and two surviving vertices will be adjacent if and only if they were adjacent before the reduction.  This last point  may induce the creation of disconnected reduced graphs. Moreover, the drop of non-survivors' features leads to an important loss of information. Let us note that MVPool~\cite{zhang2021hierarchical} increases the density of the graph by considering power $2$ or $3$ of the adjacency matrix in order to limit the disconnections of the reduced graph. It additionally adds edges to the reduced graph through an additional layer called Structure Learning.
        
        An alternative solution consists to drop Equation~\eqref{eq:3} and to perform a Kron reduction~\cite{bianchi2020hierarchical} in order to connect all pairs of  surviving vertices adjacent to a same  removed vertex.  This reduction increases the density of the graph and a sparsification step is required. This last point can creates a disconnected reduced graph. Moreover, the time complexity of the Kron reduction is approximately $\mathcal{O}((\frac{n_{l}}{2})^{3})$, due to the inversion of a part of the Laplacian matrix of $\mathcal{G}^{(l)}$.
    
        Let us finally note that it exists a graph pooling method which uses an approximation of a Maximum Independent Vertex Set called MEWIS~\cite{nour2021max}. The layer used to compute this approximation significantly increases the complexity of the whole GNN. Moreover, the approximated result provided by this layer does not guarantee that the resulting set of selected vertices is even maximal.
        
        Unlike other methods, we propose to preserve the structure of the original graph as well as the attribute information. We satisfy these properties thanks to a selection of surviving vertices distributed equally on the graph and with an assignment of non-surviving vertices to surviving vertices. We named our pooling method MIVSPool.
    
    \section{Proposed Method}
        \label{sec:proposedMethod}
        \subsection{Maximal Independent Vertex Set (MIVS)}
        Before describing the algorithm, we introduce the notion of Maximal Independent Set and show how this notion can be applied to graph vertices.
        
        \textbf{Maximal Independent Set.}
        Let $\mathcal{X}$ be a finite set and $\mathcal{N}$ a neighborhood function. A subset $\mathcal{J}$ of $\mathcal{X}$ is independent if:
        \begin{equation}
            \forall (x,y) \in \mathcal{J}^{2} : x \notin \mathcal{N}(y)\label{eq:6}
        \end{equation}
        $\mathcal{J}$ is a subset of $\mathcal{X}$ such that for any $(x,y) \in \mathcal{J}^{2}$, $x$ and $y$ are not neighbors. The elements of $\mathcal{J}$ are called the surviving elements.
        
        An independent set $\mathcal{J}$ is said to be maximal when no element can be added to it without breaking the independence property, i.e., when we have:
        \begin{equation}
            \forall x \in \mathcal{X}-\mathcal{J}, \exists y \in \mathcal{J} : x \in \mathcal{N}(y)\label{eq:7}
        \end{equation}
        Equation~\eqref{eq:7} states that each non-surviving element has to be in the neighborhood of at least one element of $\mathcal{J}$. The elements of $\mathcal{J}$ are denoted as survivors.
        
        Using Equation~\eqref{eq:6} and \eqref{eq:7}, we have a Maximal Independent Set. We note that it is a maximal but not necessarily a  maximum. Indeed, our Maximal Independent Set $\mathcal{J}$ is not necessarily those whose cardinality is maximum with respect to all Maximal Independent Sets of $\mathcal{X}$.
        
        If we interpret the construction of a Maximal Independent Set as a subsampling operation, Equation~\eqref{eq:6} can be interpreted as a condition preventing the oversampling (two adjacent vertices cannot be simultaneously selected) which thus guarantees an uniform distribution of survivors. Conversely, Equation~\eqref{eq:7} prevents subsampling: Any non-selected vertex is at a distance 1 from a surviving vertex.
        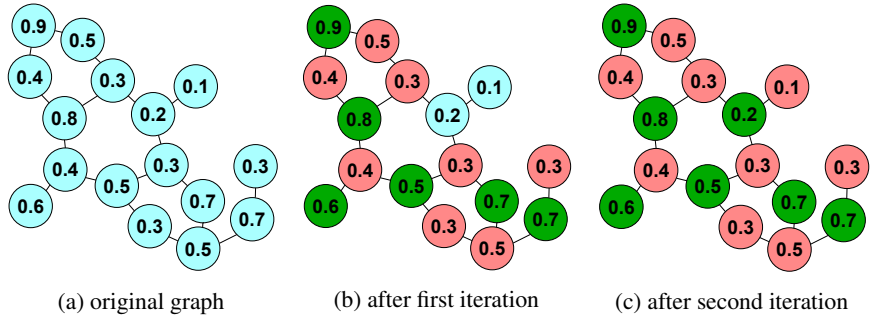
\begin{figure}[t]
            \centering
            \mbox{ } \hfill
            \begin{subfigure}{.3\textwidth}
            \centering
            \includegraphics[width=\textwidth]{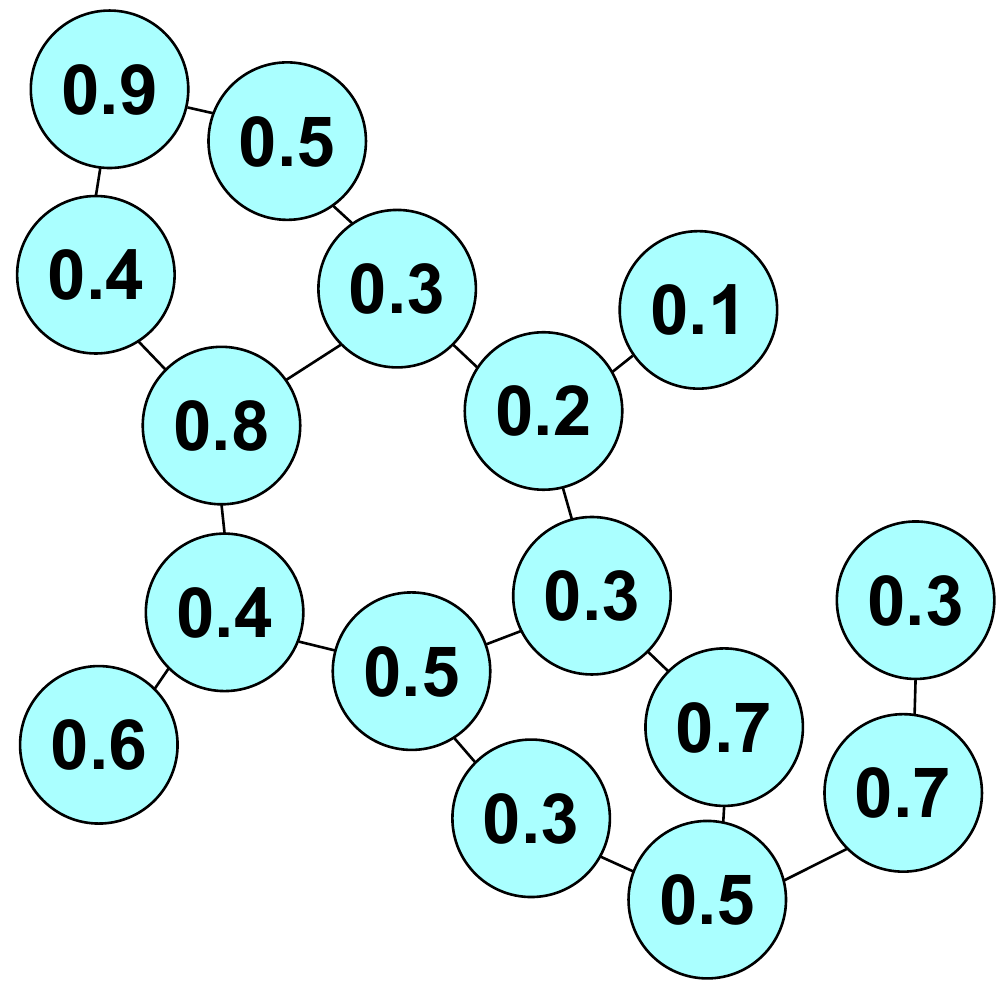}
            \caption{original graph}
            \end{subfigure}
            \hfill
            \begin{subfigure}{.3\textwidth}
            \centering
            \includegraphics[width=\textwidth]{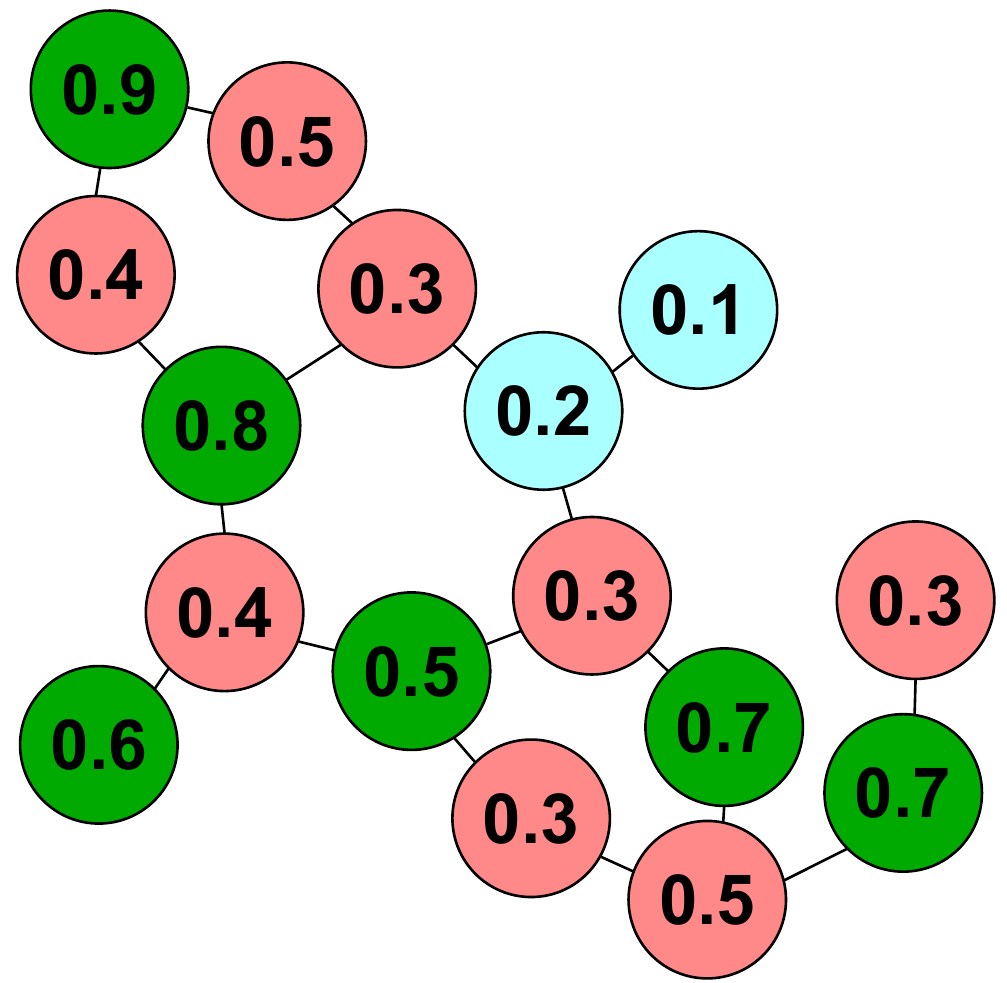}
            \caption{after first iteration}
            \end{subfigure}
            \hfill
            \begin{subfigure}{.3\textwidth}
            \centering
            \includegraphics[width=\textwidth]{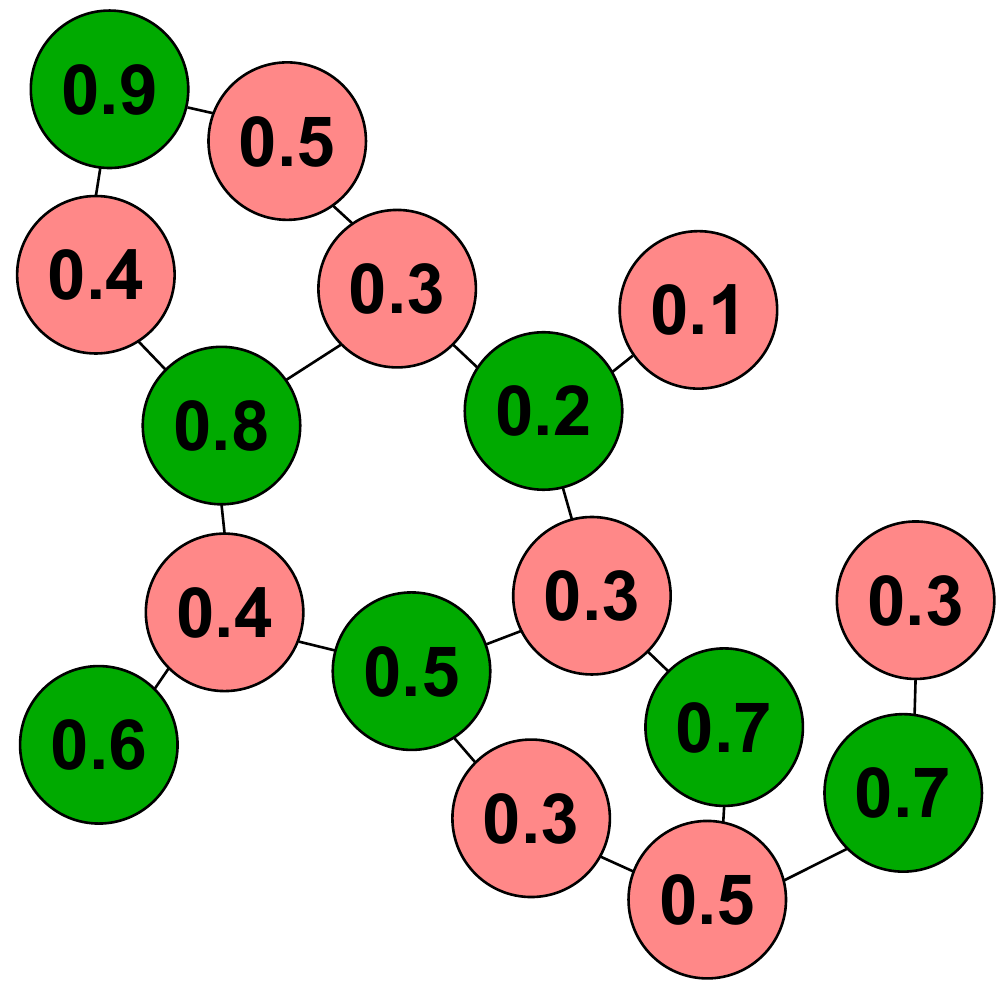}
            \caption{after second iteration}
            \end{subfigure}
            \hfill\mbox{ }
            \caption{Evolution of MIVS algorithm on a graph from the NCI1 dataset (Section~\protect\ref{subsec:datasets}). Number inside each vertex corresponds to its score $s$.  Candidate, Survivor and Non-Survivor vertices are respectively denoted by: \protect\color[HTML]{aaffff}\begin{picture}(5,5)\put(3,3){\circle*{7}}\color{black}\put(3,3){\circle{7}}\end{picture} 
            \color{black},
            \protect\color[HTML]{00aa00}\begin{picture}(5,5)\put(3,3){\circle*{7}}\color{black}\put(3,3){\circle{7}}\end{picture}
            \color{black},
            \protect\color[HTML]{FF8888}\begin{picture}(5,5)\put(3,3){\circle*{7}}\color{black}\put(3,3){\circle{7}}\end{picture}
            \color{black}.
            }
            \label{fig:exemple MIVS}
        \end{figure}        
        
        \textbf{Maximal Independent Vertex Set.}
        A Maximal Independent Vertex Set (MIVS)~\cite{meer1989stochastic} of a graph $\mathcal{G}^{(l)} = (\mathcal{V}^{(l)}, \mathcal{E}^{(l)})$ is therefore a Maximal Independent Set where the neighborhood is deduced from the edge set $\mathcal{E}^{(l)}$. Adapting Equation \eqref{eq:6} and \eqref{eq:7}, we select surviving vertices $\mathcal{V}^{(l+1)}$ as described below:
        \begin{equation}
            \forall (v,v') \in (\mathcal{V}^{(l+1)})^{2} : (v,v') \notin \mathcal{E}^{(l)}\label{eq:8}
        \end{equation}
        \begin{equation}
            \forall v \in \mathcal{V}^{(l)} - \mathcal{V}^{(l+1)}, \exists v' \in \mathcal{V}^{(l+1)} : (v,v') \in \mathcal{E}^{(l)}\label{eq:9}
        \end{equation}
        Equations~\eqref{eq:8} and \eqref{eq:9} define the MIVS procedure and state that two surviving vertices cannot be neighbors and a non-surviving vertex must have at least one survivor in its neighborhood. Nevertheless, these two equations don't explain how to select these vertices.
       
        \textbf{Meer's Algorithm.}
        A simple method has been proposed by Meer~\cite{meer1989stochastic} using an iterative procedure assigning a uniform random variable $\mathcal{U}([0, 1])$ to each vertex. In this procedure, each surviving vertex corresponds at a local maximum of the random variable with respect to its neighbors. According to Equation~\eqref{eq:8}, vertices adjacent to these survivors are labeled as non-survivors. Other vertices (not yet labeled) are labeled candidates and the algorithm iterates the selection of surviving and non-surviving vertices on this reduced vertex set (Fig.~\ref{fig:exemple MIVS}). The algorithm convergence is guaranteed since, at each iteration, at least one candidate is labeled as a survivor.
        For each vertex $v_i$, its random variable is denoted by $x_i$ while Booleans  $p_i$ and $q_i$ are $True$ if $v_i$ is respectively  a survivor and  a candidate. The stopping criterion of the algorithm is obtained when all $q_{i}$ are $False$, i.e. when Equation~\eqref{eq:9} is satisfied.      At the initialisation of the algorithm, all $p_{i}$ are $False$ and all $q_{i}$ are $True$ and, for each iteration $k$, variables $p^{(k)}_{i}$ and $q^{(k)}_{i}$ are updated by:
        
        \begin{equation}
            \begin{split}
                p^{(k+1)}_{i} = &\ p^{(k)}_{i} \lor (q^{(k)}_{i} \wedge (x_{i} = \max\{{x_{j} | (v_{i},v_{j}) \in \mathcal{E}^{(l)} \land q^{(k)}_{j}}\}))\\
                q^{(k+1)}_{i} = &\ \land_{j | (v_{i},v_{j}) \in \mathcal{E}^{(l)}} \overline{p}^{(k+1)}_{j}
            \end{split}
            \label{eq:10}
        \end{equation}
        In other words, a survivor at iteration $k+1$ is a survivor at iteration $k$ or is a candidate whose random variable ($x_i$) is greater than the ones of its candidate neighbors. A vertex is candidate at iteration $k+1$ if it is not adjacent to a survivor. We note that the neighborhood includes the central vertex. This procedure only involves local computations and is therefore parallelizable.
        
        \subsection{Adaptation of MIVS to deep learning}
        \textbf{Scoring system.}
        \label{subsec:scoring}
        Using Meer's algorithm~\cite{meer1989stochastic}, the uniform random variable $\mathcal{U}([0, 1])$ plays an important role in the selection of the surviving vertex set. We propose to modify this variable so that $x_{i}$ used in the algorithm is learnt and represents the relevance of vertex $v_{i}$. We obtain this last property by using an attention mechanism like in SagPool~\cite{lee2019self} where a score vector $s \in \mathbb{R}^{n_l \times 1}$ is returned by a GCN~\cite{kipf2017semi}. Using this score in our MIVS, we select a set of surviving vertices $\mathcal{V}^{(l+1)}$ corresponding to local maxima of the function of interest encoded by $s$. An uniform distribution of vertices on the graph is guaranteed by the computation of the MIVS.
        
        \textbf{Assignment of Non-Surviving Vertices.}
        In order to construct our reduction matrix $S^{(l)}$ and take all vertices information into account, we need to assign non-surviving vertices. As a reminder, Equation~\eqref{eq:9} states that each non-survivor has at least a surviving neighbor. Assuming that the score function encodes the relevance of each vertex, we assign each non-surviving vertex to its  surviving neighbor with the highest score.  We obtain a reduction matrix $S^{(l)}$ with $n_{l+1}$ clusters corresponding to surviving vertices:
        \begin{equation}
            \begin{split}
                \forall v_{j} \in \mathcal{V}^{(l)} - \mathcal{V}^{(l+1)}, \exists! v_{i} \in \mathcal{V}^{(l+1)} | \mathcal{S}^{(l)}_{ji} = 1\\
                \text{with } i = argmax(s_k, (v_{k}, v_{j}) \in \mathcal{E}^{(l)} \land p^{N}_{k})
            \end{split}\label{eq:11}
        \end{equation}
        where N is the number of iterations required to have a MIVS and $s$ the score vector.
        
        \textbf{Construction of reduced attributes and adjacency matrix.}
        The construction of attributes of the reduced graph $\mathcal{G}^{(l+1)}$ is obtained thanks to an average weighted by $s$ from the set of aggregated vertices to the surviving vertices:
        \begin{equation}
            X^{(l+1)}_{i} = \frac{1}{\sum_{j | S^{(l)}_{ji}=1} s_{j}} \sum_{j | S^{(l)}_{ji}=1} s_{j} X^{(l)}_{j}
            \label{eq:12}
        \end{equation}
        Equation~\eqref{eq:12} allows to take into account the importance of each vertex in the computation of the attributes of the reduced graph and put more attention on vertices with a high score. As a consequence the learnt vector $s$ can be interpreted as a relevance value. This operation can be achieved by a transformation similar to  Equation~\eqref{eq:2} by substituting $S^{(l)}$ by the matrix $S^{(l)'}~=~D_{2}~D_{1}~S^{(l)}$ with $D_{1}=diag(s)$ and $D_{2}=diag(\frac{1}{\mathbf{1}\top D_{1}S^{(l)}})$.
        The construction of the reduced adjacency matrix $A^{(l+1)}$ is obtained thanks to  Equation~\eqref{eq:3}.
        
         \textbf{Relaxation of MIVSPool.} 
         On some highly dense graphs, MIVSPool may provide a decimation ratio lower than $0.5$. In order to correct this point, we additionally introduce MIVSPool$_{comp.}$ which is a MIVSPool where we force the addition of surviving vertices using a Top-\textit{k} so that the pooling ratio remains always equal to $0.5$. Note that Equation~\eqref{eq:8} is then no more valid while Equation~\eqref{eq:9} still holds.
    \section{Experiments}
        \label{sec:xp}
        \subsection{Datasets}
        \label{subsec:datasets}
         \begin{table}[b!]
            \centering
            \begin{tabularx}{.8\textwidth}{lCCCC}\toprule
                \textbf{Dataset} & \textbf{\#Graphs} & \textbf{\#Classes} & \textbf{Avg $|\mathcal{V}|$} & \textbf{Avg $|\mathcal{E}|$} \\
                \hline
                D\&D & $1178$ & $2$ & $284 \pm 272$ & $715 \pm 694$ \\
                PROTEINS & $1113$ & $2$ & $39 \pm 46$ & $72 \pm 84$ \\
                NCI1 & $4110$ & $2$ & $29 \pm 13$ & $32 \pm 14$ \\
                ENZYMES & $600$ & $6$ & $33 \pm 15$ & $62 \pm 26$ \\
                \bottomrule
            \end{tabularx}
            \caption{Statistics of datasets}
            \label{tab:dataset}
        \end{table}   
        \begin{table}[b!]
                \centering
                \begin{tabularx}{\textwidth}{lCCCC}\toprule
                    \textbf{Score function} & \textbf{MIVSPool$_{rand}$} & \textbf{MIVSPool$_{Top-\textit{k}}$} & \textbf{MIVSPool$_{SagPool}$} & \textbf{MIVSPool$_{MVPool}$}\\
                    \hline
                    D\&D    &   $77.04 \pm 0.63$ & $ 77.10 \pm 1.00$ & $75.10 \pm 0.76$ & $\mathbf{77.38 \pm 0.94}$ \\
                    PROTEINS&   $75.15 \pm 0.44$ & $75.36 \pm 0.60$ & $75.11 \pm 0.74$ & $\mathbf{75.62 \pm 0.47}$ \\
                    NCI1    &   $72.16 \pm 0.55$ & $\mathbf{73.82 \pm 0.94}$ & $73.72 \pm 0.71$ & $72.97 \pm 0.71$ \\
                    ENZYMES    &    $45.80 \pm 1.35$ & $37.55 \pm 1.94$ & $38.68 \pm 2.81$ & $\mathbf{46.80 \pm 1.53}$ \\
                    \bottomrule
                \end{tabularx}
                \caption{Study of MIVSPool according to the score function}
                \label{tab:score_function}
        \end{table}
        To evaluate our MIVSPool, we test it on a benchmark of four standard datasets: D\&D~\cite{dobson2003distinguishing}, PROTEINS~\cite{BorgwardtOSVSK05,dobson2003distinguishing}, NCI1~\cite{wale2008comparison} and ENZYMES~\cite{BorgwardtOSVSK05}. The statistics of datasets are reported on Table~\ref{tab:dataset}. D\&D and PROTEINS describe proteins and the aim is to classify them as enzyme or non-enzyme. Nodes represent the amino acids and two nodes are connected by an edge if they are less than 6 \AA{}ngström apart. NCI1 describes molecules and the purpose is to classify them as cancerous or non-cancerous. Each vertex stands for an atom and edges between vertices represent bonds between atoms. ENZYMES is a dataset of protein tertiary structures obtained from the BRENDA enzyme database. 
        
        \subsection{Model Architecture and Training Procedure}
        The model architecture consists of K blocks made up of a GCN~\cite{kipf2017semi} convolution followed by a graph pooling. A Readout layer is applied after each block using a concatenation of the average and the maximum of vertices' features matrix $X^{(l)}$. At the end of our network, the K Readout are summed and the result is sent to a Multilayer Perceptron. The latter is composed by three fully connected layers (respectively 256, 128 and 64 for the number of hidden neurons)  and a droupout of $0.5$ is applied between the first two. The classification is obtained by a Softmax layer (see Fig.~\ref{fig:model}).
        
        For the training procedure, we use Pytorch Geometric and we evaluate our neural network with a 10-fold cross validation. We repeat this procedure ten times without setting the seed. The dataset is split in three parts: 80\% for the training set, 10\% for the validation set and 10\% for the test set.
        
        For the hyperparameters, we use the Adam optimizer, set the dimension of node representation, the batch size and the number of epochs respectively at 128, 512 and 1000 and an early stopping is applied if the validation loss did not improved after 100 epochs. A grid search is used for the learning rate within $\left\{1e^{-3},1e^{-4},1e^{-5}\right\}$, the weight decay within $\left\{1e^{-3},1e^{-4},1e^{-5}\right\}$ and the number of blocks K in $\left[3,5\right]$ to find the best configuration.
        \begin{table}[b!]
                \centering
                \begin{tabularx}{0.85\textwidth}{lCCCCC}\toprule
                    \textbf{Dataset} & \textbf{Pooling 1} & \textbf{Pooling 2} & \textbf{Pooling 3} & \textbf{Pooling 4} & \textbf{Pooling 5} \\
                    \hline
                    D\&D &  $4.1 \pm 0.6$  &  $3.6 \pm 0.6$  &  $3.2 \pm 0.6$  & - & - \\
                    PROTEINS &  $3.1 \pm 0.7$  &  $2.7 \pm 0.6$  &  $2.3 \pm 0.5$  & - & - \\
                    NCI1 &  $3.2 \pm 0.5$  &  $2.9 \pm 0.5$  &  $2.5 \pm 0.5$  &  $2.3 \pm 0.5$ & $2.1 \pm 0.3$   \\
                    ENZYMES &  $3.2 \pm 0.6$  &  $2.7 \pm 0.6$  &  $2.4 \pm 0.5$  & - & - \\
                    \bottomrule
                \end{tabularx}
                \caption{Averaged number of iteration of MIVS for each pooling step.}
                \label{tab:nb_iter}
            \end{table}
        \subsection{Ablation Studies}
            \textbf{Scoring function.}
             encodes the relevance of each vertex and as a direct influence on the set of selected surviving vertices by MIVSPool. We vary it using  a uniform random variable $\mathcal{U}([0, 1])$ like in Meer~\cite{meer1989stochastic} ( MIVSPool$_{rand}$),  a trainable normalized projection vector on the features $X^{(l)}$ like in~\cite{gao2019graph} (MIVSPool$_{Top-\textit{k}}$),  a self-attention mechanism thanks to a graph convolution like in SagPool~\cite{lee2019self} (MIVSPool$_{SagPool}$) and finally a trainable multi-view system across structure and features graph information like in MVPool~\cite{zhang2021hierarchical} (MIVSPool$_{MVPool}$).
            For trainable variations, Equation~\eqref{eq:12} allows to propagate the training to the next convolution and, therefore, our pooling method is end-to-end trainable. The results reported Table~\ref{tab:score_function} show that the choice of the score function has a minor influence on the accuracy. Note that, since Top-\textit{k} and SagPool trainable score functions don't consider structural information, they consequently perform poorly on ENZYMES. However, the use of the MVPool trainable score function allows to obtain the best accuracy on three datasets with comparable  standard deviations  to other heuristics. In the rest of our article, we choose to take MIVSPool$_{MVPool}$ as a reference that we denote MIVSPool.%

            \textbf{Average number of iterations required by  MIVSPool for each pooling step.}
            Table~\ref{tab:nb_iter} presents the mean number of iterations for each dataset and each pooling step computed over 10 epochs. Note that the number of pooling steps is determined using K-folds for each dataset.
            Despite an important difference between  graph sizes among datasets~(Table~\ref{tab:dataset}), we note that the number of iterations is comparable between them and less than $5$. Knowing that an iteration of MIVS on a graph $\mathcal{G}=(\mathcal{V},\mathcal{E})$ requires about $|\mathcal{V}|d_{max}$ computations, where $d_{max}$ is the maximum degree of  $\mathcal{G}$, the computation of MIVS on one of the graphs of our three datasets is bounded by $5|\mathcal{V}|d_{max}$. This complexity is less than the one needed by the Kron transformation (Section~\ref{subsec:state_art_pooling}).
         
        \subsection{Comparison of MIVSPool according to other methods}
        We compare  our method to  three state-of-art methods: gPool~\cite{gao2019graph}, SagPool~\cite{lee2019self} and MVPool-SL~\cite{zhang2021hierarchical}.  Results   in Table~\ref{tab:comparison_other_methods}  show that MIVSPool$_{comp.}$ and MIVSPool obtain  the highest or second highest accuracies on D\&D, PROTEINS and ENZYMES. This point confirms the efficiency of MIVSPool and the fact that  some configurations (highly connected graphs) may induce low decimation rate by MIVSPool which slightly decreases the accuracy. For NCI1, the highest accuracy is obtained by MVPool-SL~\cite{zhang2021hierarchical},  MIVSPool$_{comp.}$ being ranked  second. Let us note that the mean vertex's degree in this dataset is $2.15\pm 0.11$ with  many graphs being almost linear. On such simplified topology considering two hops has done by MVPool-SL allows to recover the structure of the reduced graph.
        \begin{table}[b]
            \centering
            \begin{tabularx}{\textwidth}{lCCcCC}\toprule
                \textbf{Dataset}
                & \textbf{gPool~\cite{gao2019graph}} & \textbf{SagPool~\cite{lee2019self}} & \textbf{MVPool-SL~\cite{zhang2021hierarchical}} & \textbf{MIVSPool} & \textbf{MIVSPool$_{comp.}$}\\
                \hline
                D\&D & 
                $75.76 \pm 0.82$  &  $75.92 \pm 0.92$  &  $77.26 \pm 0.37$  &  \textcolor{blue}{$77.38 \pm 0.94$} &  $\mathbf{77.88 \pm 0.73}$  \\
                PROTEINS & 
                 $73.49 \pm 1.44$  &  $74.30 \pm 0.62$  &  $75.04 \pm 0.67$  &  \textcolor{blue}{$75.62 \pm 0.47$} &  $\mathbf{75.81 \pm 0.75}$ \\
                NCI1 &
                $71.66 \pm 1.04$  &  $72.86 \pm 0.68$  &$\mathbf{74.79 \pm 0.58}$  &  $72.97 \pm 0.71$  &  \textcolor{blue}{$73.44 \pm 0.68$} \\
                ENZYMES &
                $36.93 \pm 2.45$  & $35.30 \pm 1.80$  & $39.45 \pm 2.57$  &  $\mathbf{46.80 \pm 1.53}$  &  \textcolor{blue}{$45.60 \pm 2.37$} \\
                \bottomrule
            \end{tabularx}
            \caption{Comparison of MIVSPool according to other hierarchical methods. Highest and second highest accuracies are respectively in \textbf{bold} and \textcolor{blue}{blue}.}
            \label{tab:comparison_other_methods}
        \end{table}
    \section{Conclusion}
    Our graph pooling method MIVSPool is based on a selection of surviving vertices thanks to a Maximal Independent Vertex Set (MIVS) and an assignment of non-surviving vertices to surviving ones. Unlike state-of-art methods, our method allows to preserve the totality of graph information during its reduction. 
    \paragraph{Acknowledgements:}
    The work was performed using computing resources of CRIANN (Normandy, France).

%
%
%
    \bibliographystyle{splncs04}
    \bibliography{bibliography}

\end{document}